%% file: 00-aclwkp2023-4page.tex
\pdfoutput=1

\documentclass[11pt]{article}

\usepackage{ACL2023}

\usepackage{times}
\usepackage{latexsym}

\usepackage[T1]{fontenc}

\usepackage[utf8]{inputenc}

\usepackage{microtype}

\usepackage{inconsolata}
\usepackage{multirow}
\usepackage{algorithm}
\usepackage{algorithmic}

\input{preambular}

\title{Expanding Scope: Adapting English Adversarial Attacks to  Chinese}

\author{
  Hanyu Liu, Chengyuan Cai, Yanjun Qi \\
  Department of Computer Science \\
  University of Virginia \\
  Charlottesville, VA, USA \\
  \texttt{\{hl2gn, cc4fy, yq2h\}@virginia.edu} \\
  }

\begin{document}
\maketitle
\begin{abstract}

Recent studies have revealed that NLP predictive models are vulnerable to adversarial attacks. Most existing studies focused on designing attacks to evaluate the robustness of NLP models in the English language alone. Literature has seen an increasing need for NLP solutions for other languages. We, therefore, ask one natural question: whether state-of-the-art (SOTA) attack methods generalize to other languages. This paper investigates how to adapt SOTA adversarial attack algorithms in English to the Chinese language. Our experiments show that attack methods previously applied to English NLP can generate high-quality adversarial examples in Chinese when combined with proper text segmentation and linguistic constraints. In addition, we demonstrate that the generated adversarial examples can achieve high fluency and semantic consistency by focusing on the Chinese language's morphology and phonology, which in turn can be used to improve the adversarial robustness of Chinese NLP models. 

\end{abstract}

\input{01-intro}

\section{Method}

\input{02-back}

\input{10-method}

\input{11-details-transform}

\input{12-constraints}

\input{33-figure}

\section{Results and Evaluation}
\input{20-setup}
\input{21-results}

\input{22-at-short}

\bibliography{ref}
\bibliographystyle{acl_natbib}

\section{Limitations}

We are optimistic that the algorithmic workflow presented in this paper can be generalized to other languages. When the victim models are in languages other than Chinese and English, however, we also acknowledge the uncertainty in achieving a high attack success rate while at the same time achieving fluency in generated examples. In addition, because of the variation in linguistic structures across different languages, further efforts are required to design language-specific transformation methods (such as the homophone and morphonym transformations for the Chinese language case in this paper).

\section{Ethics Statement}

In this study, we honor the ethical code in the ACL Code of Ethics.

\appendix

\input{02-more-back}

\section{More Method Details}
\input{11-details-search}

\input{12-more-c}

\section{More on Results and Setup}
\label{sec:more:exp}
\input{30-tables}

\input{22-at-results}
\input{20-more-data}

\input{23-discussion}

\input{25-conclusion}

\section{Qualitative Examples}
\label{sec:figures}

\input{32-figure}

\input{31-figure}

\newpage

\end{document}

%% file: preambular.tex
\usepackage{amsmath}
\usepackage{amsfonts}
\usepackage{amssymb}

\usepackage{caption}
\usepackage{subcaption}

\newcommand{\importanceRankingName}{word importance ranking }

\newcommand{\ignore}[1]{}

\usepackage{color}

\newcommand{\tref}[1]{Table~\ref{#1}} 
\newcommand{\fref}[1]{Figure~\ref{#1}} 
\newcommand{\sref}[1]{Section~\ref{#1}} 
\newcommand{\eref}[1]{Eq.~(\ref{#1})}

\newcommand{\cref}[1]{Condition~(\ref{#1})}

\usepackage{url}
\usepackage{wrapfig}
\usepackage{enumitem}

\def\x{{\mathbf x}}

\def\F{{\mathbf F}}

\usepackage{adjustbox}

\graphicspath{{../figs/}{../figures/}{../Resources/}{Resources/}}

\usepackage{booktabs}

\usepackage{hyperref}
\usepackage{url}
\usepackage{dirtytalk}

\usepackage{graphicx} 
\usepackage{float} 
\usepackage{amsmath}
\usepackage{color,xcolor}
\usepackage{array}
\usepackage{xspace}
\usepackage{multirow}
\usepackage{csquotes}
\usepackage{booktabs}

\usepackage{url}
\usepackage{multirow}
\usepackage{graphicx} %
\graphicspath{{fig/}} %

\usepackage{enumitem}
\setlist{nosep}

\usepackage{lipsum}
\usepackage{titlesec}

 \titlespacing\section{0pt}{3pt plus 0pt minus 2pt}{1pt plus 1pt minus 1pt}
 \titlespacing\subsection{0pt}{2pt plus 0pt minus 2pt}{1pt plus 1pt minus 1pt}
 \titlespacing\subsubsection{0pt}{2pt plus 0pt minus 2pt}{1pt plus 1pt minus 1pt}
 \titlespacing{\paragraph}{0pt}{2pt plus 0pt minus 1pt}{1pt plus 1pt minus 0pt}[2pt]

\usepackage[font=footnotesize]{caption}

\usepackage{etoolbox}
\BeforeBeginEnvironment{figure}{\vskip-1ex}
\AfterEndEnvironment{figure}{\vskip-0.2ex}

\usepackage{caption}
\allowdisplaybreaks
\newcommand{\shorten}[1]{}

%% file: 01-intro.tex
\section{Introduction}

Adversarial examples are text inputs crafted to fool an NLP system, typically by making small perturbations to a seed input\footnote{Most existing work attempts to perturb an input  using character-level  \citep{ebrahimi2017hotflip, Gao2018BlackBoxGO, pruthi2019combating, li2018textbugger} or word-level perturbations \citep{alzantot2018generating, Jin2019TextFooler, pwws-ren-etal-2019-generating, pso-zang-etal-2020-word} to fool a target model's prediction in a specific way.}.  Recent literature has developed various adversarial attacks generating text adversarial examples to fool NLP predictive models \footnote{We use ``natural language adversarial example'', ``text adversarial example'' and "adversarial attacks" interchangeably.}. These attack methods mainly focus on the English language alone, building upon components that use language-specific resources, such as English WordNet \citep{wordnet} or BERT models \citep{devlin2018bert} pretrained on English corpus.

Literature has seen a growing need for NLP solutions in other languages; therefore, evaluating NLP solutions' robustness via adversarial examples is crucial. We ask an immediate question: "Can we extend the SOTA adversarial attacks in English to other languages by replacing those English-specific inner components with other languages' resources ?". For instance, we can attack a Chinese NLP model by replacing WordNet with HowNet \citep{HowNet-2010}. However, it is unclear if such a workflow is sufficient for generating high-quality adversarial examples, when a target language differs from English. In this work, we attempt to answer this question by adapting SOTA word substitution attacks designed for English to evaluate Chinese NLP models' adversarial robustness. Moreover, we introduce morphonym and homophone word-substitution attacks that are specific to the Chinese language; they function as a benchmark to the English adapted attack methods.

Our experiments on Chinese classification and entailment models show that both the English-adapted and Chinese-specific attack methods can effectively generate adversarial examples with good readability. The attack success rates of homophone-based and HowNet-derived methods are significantly better than the success rate of masked language model-based attacks or morphonym-derived attacks. We then combine the four attacks mentioned above into a composite attack that further increases the attack success rate to 96.00\% in fooling Chinese classification models and 98.16\% in attacking entailment models. In addition, we demonstrate that adversarially trained models significantly decrease attack success rate by up to 49.32$\%$.

%% file: 02-back.tex
Recent NLP literature includes a growing body of works on adversarial examples in NLP, mostly in English (more background details are in \sref{back:ae}). Most SOTA English adversarial attacks search for a perturbation to change a given seed input $\x$ into an adversarial example $\x'$;  $\x'$ fools a predictive NLP model and satisfies certain language constraints, like preserving the same semantical meaning as $\x$. Essentially each adversarial attack algorithm has four components: a goal function, a set of constraints, a suite of transformations, and a search algorithm \cite{morris2020reevaluating}. The search algorithm attempts to find a sequence of transformations that results in a successful perturbation. The goal function can be like fooling a target model into predicting the wrong classification label.

\paragraph{Related literature:}  While most NLP adversarial attacks have focused on the English language, a few recent methods have been proposed for Chinese. \citet{argot2020Chinese} proposed a black-box attack that performs a glyph-level transformation on the Chinese characters. Related, \citet{textshield} and \citet{zhang-etal-2022-rochbert} added phonetic perturbations to improve the adversarial robustness of Chinese NLP models. All three attacks, however, are only applicable to the Chinese language. Another study \cite{wang2020evaluatingChineseBERT}  proposed a white-box attack against BERT models \cite{Devlin2018BERTPO} that performs character-level swaps using gradient optimization. These character-level attacks extend poorly to other languages and tend to generate out-of-context partial substitutions that impact fluency. Later studies, such as \citet{GPSAttack} and \citet{wang-etal-2022-semattack},  included semantic-based word substitutions but did not consider the significance of constraints and adversarial training. We choose to generalize SOTA word synonym substitution attacks in English to the Chinese language (due to the prevalence of word substitutions) and our attacks consider a range of language constraints.

\subsection{Determining Text Segmentation}

The first step to crafting a new adversarial attack for the Chinese language is to select the level of transformation. Unlike English, which separates words with space, the Chinese language lacks native separators to determine different words in a sentence. A Chinese character may represent a word, while longer words may include multiple adjacent Chinese characters. To avoid out-of-context perturbations that replace partial components of a multi-character word, we use a Chinese segmentation tool provided by Jieba \footnote{\url{https://github.com/fxsjy/jieba}} to segment an input text into a list of words.

%% file: 10-method.tex
\subsection{General Overview of Proposed Attacks}
\label{sec:chinese}

The general perturbation strategy we propose is word synonym substitutions. Given an input text $\x$, we use the aforementioned segmentation tool to segment $\x$ into $[x_1, x_2, \dots, x_n]$. 
Subsequent transformations (synonym substitution) are then getting applied to each eligible word \footnote{In this paper, the phrase "Chinese characters" refers to one unit long token, and "Chinese words" refers to one or more Chinese characters in their semantically correct segmentation that may or may not be one unit long.}. This means we obtain perturbed text $\x'$ by replacing some $x_i$ with its synonym $x'_i$. Our attack goal is to make the model mis-predict the $\x'$  (i.e. $\F(\x) \neq \F(\x')$), \footnote{Here  $\F:\mathcal{X}\rightarrow \mathcal{Y}$ denotes a predictive Chinese NLP model taking Chinese text as input. $\mathcal{X}$ denotes the input space and $\mathcal{Y}$ is the output space.} which is also called an untargeted attack.  If one substitution is not enough to change the prediction, we repeat the steps to swap another $x_j$ to generate the perturbed text $\x'$.
This process essentially solves the following objective: 
\begin{equation}
  \begin{split}
    \label{eq:adv}
    \text{Find } & \ \ \x' = \text{ wordSubstitution} (\x) \\ 
    \text{s.t. } & \ \  \F (\x) \neq \F(\x') \\
    &  \ \ \x' \in \mathcal{X},\:  \F (\x) = y_{orig}\\
            & \wedge C_i(\x, \x'; \epsilon_i),   \: \forall i \in \{1, 2, ..., C\} 
  \end{split}
\end{equation}
Here ${C_1,...,C_n}$ denotes a set of language constraints including like semantic preserving and grammaticality constraints \cite{morris2020reevaluating}. $\epsilon_i$ denotes the strength of the constraint $C_i$.

The critical component "$ \text{ wordSubstitution} (\x) $" in \eref{eq:adv} requires us to figure out what words in $\x$ to perturb first, and what words as next. Essentially this is a combinatorial search issue. Literature includes different search strategy (see \sref{ref:search} for details). We adapt the greedy with \importanceRankingName based search algorithm here. 
Our attack chooses the order of words by estimating the ``importance" of each $x_i$ in $\x$. The importance of $x_i$ is computed by replacing each $x_i$ with an \texttt{UNK} token and then calculating the change in the model's confidence on the original label. Essentially we sort words $x_i$ in $\x$ by the decreasing importance regarding the following  $score$:
\begin{equation}
  \begin{split}
    \label{eq:score}
    score(x_i) & = 1 - Prob(\F(\x'))_{y_{orig}}  \\
    \textbf{s.t.      } \    \ & \x'=  \text{replace} (\x, x_i, \text{UNK})
  \end{split}
\end{equation}
This measures how much the target model's confidence decreases regarding the original label class $y_{orig}$ when replacing $x_i$ with "UNK" token. Then for each selected $x_i$, we find its best $x_i'$ to swap with, from a candidate synonym set (\sref{sec:synonym})

\subsection{Generating Synonyms for Words}
\label{sec:synonym}

Now for a selected word $x_i$ in $\x$,  we propose four different Chinese word transformation strategies to perturb a word $x_i$ into $x_i'$ through the following word transformations:

%% file: 11-details-transform.tex
We design the first two transformations by adapting from English attack studies \cite{Jin2019TextFooler} and \cite{garg-ramakrishnan-2020-bae}. 
\begin{itemize}[leftmargin=*]
 \item \textbf{Open HowNet.} Open HowNet \cite{qi2019OpenHowNet} is a sememe-based lexical dataset that is consisted of a sememe set and the corresponding phrases annotated with different sememes. A sememe is defined as the minimum semantic unit in a language, and Open HowNet incorporates relations between sememes to construct a taxonomy for each sememes. The semantic similarity between two words can be calculated by comparing their annotated sememes. In our study, we use Open HowNet to generate synonyms by searching the top five words with the highest semantic similarity with an input Chinese word. 
  \item \textbf{Masked Language Model.} We adapt the masked language model (MLM) method to generate perturbations based on the top-K predictions by a MLM. The XLM-RoBERTa model \cite{xlm-roberta} was used as the MLM in our study, as it is able to predict Chinese words consisting of multiple characters to preserve the fluency of the attacked sentence better, in comparison to other prevalent MLM (mac-bert, etc.) that predicts single characters alone.
\end{itemize}

The Chinese language, along with other Eastern Asian languages, differs from English, especially in phonology and morphology.\footnote{Each Chinese character represents a monosyllabic word with unique combinations of pictographs, while English words consist of alphabetic letters. Though each Chinese character's morphology combination is unique, many characters with similar morphology structures can be substituted in an adversarial attack without impacting the readability of the attacked sentence. In addition, because there exist many homophones in modern Chinese, the same spoken syllable may map to one of many characters with different meanings. The phonology of Chinese characters is commonly transcribed into the Latin script using Pinyin. Typing the wrong character of a word in Pinyin despite having the same pronunciation is a common mistake in Chinese writing. Thus, replacing Chinese characters with the same pronunciation may serve as an additional attack method to test the adversarial robustness of NLP models while preserving the semantics for human readers.} Using these intuitions, we design two special word transformations considering phomophones and morphonyms of Chinese language. 
\begin{itemize}[leftmargin=*]
\item \textbf{Homophone transformation.} Since the phonology of Chinese characters can be expressed by the romanization system Pinyin. To replace a Chinese character with its homophone, top-k words are randomly selected from a list of characters with the same Latin script. 
\item \textbf{Morphonym transformation.} Similarly, to replace a character with its morphonyms, top-k words are randomly selected from a list of characters that share partial pictographs with the target character, as it is a common mistake for Chinese writers to mistaken one pictograph with another.
\item \textbf{Composite transformation.} We also design a composite transformation that consists of the four transformation methods listed above. For each target word, Open HowNet, Masked Language Model, Homophone, and Morphonym perturbations are separately generated to replace a candidate word from the input text. If none of the substitutions changes the target NLP model prediction, the attack then move on to replace the next important word in the input sentence.
\end{itemize}

%% file: 12-constraints.tex
In addition, for each perturbation, we want to ensure that the generated $\x'$ preserves the semantic consistency and textual fluency of $\x$. We use three constraints, namely (1) constraint to allow only non-stop word modification, (2) constraint to allow only no-repeat modification, and (3) multilingual universal sentence encoder (MUSE) similarity constraint that filter out undesirable replacements \citep{Cer18USE} \footnote{ We require that the MUSE similarity is above $0.9$. }. These constraints can easily adapt to other languages. A detailed description of each constraint is in \sref{sec:method:morec}. The pseudo-code of our proposed attacks is in Algorithm~\ref{attack-algo}.

In summary, each word transformation strategy gets combined with the  greedy word ranking algorithm (\sref{sec:chinese}) plus the language constraints (see above), making a unique adversarial attack against Chinese NLP.

%% file: 33-figure.tex
\begin{figure*}[h!]
     \centering
     \includegraphics[width=\textwidth]{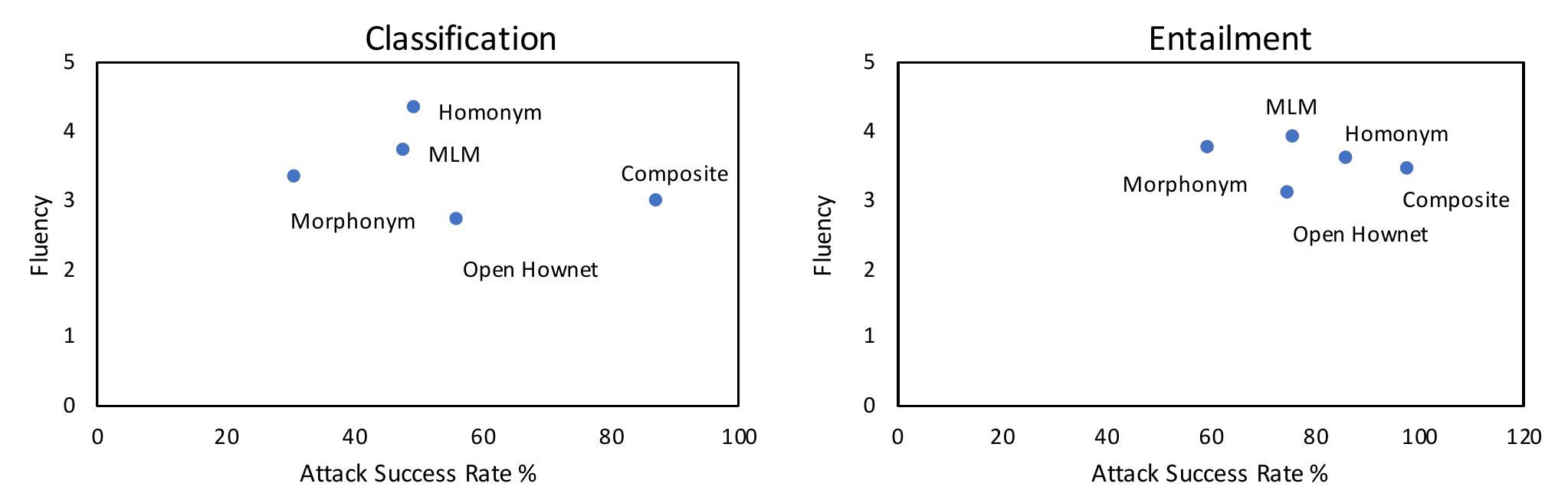}
\vspace{-5mm}
 \caption{The performance of composite attack method with  STM-RM-MUSE constraint regarding the attack success rate and human-evaluated fluency on BERT classification model (left), and RoBERTa entailment model (right). For both classification and entailment tasks, composite transformation achieves the highest attack success rate without a significant trade-off in fluency, while morphonym transformation has the lowest attack success rate.}
        \label{fig:sr_fluency}
\vspace{-5mm}
\end{figure*}

%% file: 20-setup.tex
\paragraph{Victim Models:} We chose to perform attacks on two Chinese NLP models: one for sentiment classification and one for entailment. 
BERT and RoBERTa as selected as our victim models due to their reported robustness and SOTA performance. Details of the two models and its related two Chinese datasets are presented in~\sref{sec:data}.

\paragraph{Metrics:} For each attack method, we recorded the attack success rate and perturbation percentage, skipping samples that a target model fails to predict correctly before any perturbation.

\paragraph{Ablation:} To measure how MUSE constraint impact the quality of Chinese adversarial examples, we add baseline attacks that use only the stop word constraint and repeat constraints for ablation study. 

%% file: 21-results.tex
\fref{fig:sr_fluency} connects attack success rate and fluency in one figure. \fref{fig:examples1} and \fref{fig:examples2} show few Chinese adversarial examples generated by our attacks. More results can be found in \sref{sec:discuss}

\paragraph{Results on Attack Success:} \fref{fig:sr_fluency}, \tref{tab:attack-result-class} and \tref{tab:attack-result-entail} present the quantitative results of our attacks.  \fref{fig:sr_fluency} (left) is about our results on Chinese sentiment classification model. Among all non-composite-transformation based attacks, we can see that Open HowNet substitution achieves the highest success rate, while morphonym substitution has the lowest success rate. From \tref{tab:attack-result-class}, we can also see that having the MUSE constraint dramatically decreases the attack success rate and perturbation percentage for all attack methods, especially for Open HowNet and homonym substitutions based attacks. This makes sense as the MUSE constraint is designed to limit the amount of perturbation the attacks can do to improve the quality of generated adversarial example. In addition, when we compare the success rate and perturbation percentage of composite attack versus other individual attack methods, we see that composite attack achieves a 87.50\% attack success rate without increasing the perturbation percentage. We can make similar conclusions from  \fref{fig:sr_fluency} (right) and \tref{tab:attack-result-entail}.

\shorten{
While we generally would like a high success rate, having a high success rates with examples that are not adversarial in nature is not desirable as it does not tell much about the robustness of the victim models. Because a higher perturbation percentage indicates greater deviations from the original semantics of the sentence, the decrease in perturbation percentage also implies that MUSE constraint can lead to more fluent and natural adversarial example. 
}

\paragraph{Human Evaluation:} For each of the attack method, we randomly sampled 30 adversarial examples produced from the same set of input texts for each attack (a total of five). We asked four volunteers to score the semantic consistency and fluency of the examples. Semantic consistency refers to how well the ground truth label of the adversarial example matches with the original label of the input, and fluency refers to the cohesiveness of the sentence. Both metrics are scored on a scale of 1 to 5,  with a score of 5 being the most consistent or fluent.

\tref{fig:human_eval_result_class} and \tref{fig:human_eval_result_entail}  respectively summarize the human evaluations of adversarial examples generated by fooling classification and entailment models. For classification, \tref{fig:human_eval_result_class} (plus \fref{fig:sr_fluency}) shows that homonym substitution outperforms other attack methods, as its examples have both the highest consistency and fluency scores. On the other hand, Open HowNet substitution reports the lowest quality scores, indicating its generated adversarial examples either include out-of-context substitution or disrupt the cohesiveness semantics. \tref{fig:human_eval_result_class}, plus \fref{fig:sr_fluency}(right) for entailment tasks, shows that homonym attack still achieves the highest consistency score, while MLM achieves the highest fluency score. Besides, we conjecture that the low consistency and fluency scores of the composite attack method may root to its inclusion of adversarial examples generated by Open HowNet.

%% file: 22-at-short.tex
\paragraph{Adversarial training and more result discussions:}  Furthermore, we conduct adversarial training (AT) (see details in \sref{sec:at}).  \tref{fig:at-result-class}  shows the positive results of AT that improve the robustness across all five proposed attacks over both models.

%% file: 02-more-back.tex
\section{Background: NLP Adversarial Attacks}
\label{back:ae}

Adversarial examples are inputs crafted to fool a machine learning system, typically by making small perturbations to a seed input \cite{szegedy2013intriguing, goodfellow2014explaining, papernot2016limitations, moosavi2016deepfool}. The study of natural language processing (NLP) in adversarial environments is an emerging topic as many online platforms provide NLP based information services, like toxic content detection, misinformation or fake news identification. These applications make NLP frameworks potential targets of adaptive adversaries.

Adversarial attacks aim to use a set of transformations ${T_1...T_k}$ to perturb a correctly predicted instance, $\x \in \mathcal{X}$, into an adversarial instance  $\x'$. Attacks normally define a  goal function $FoolGoal(\F, \x')$ that represents whether the goal of the attack has been met, for instance, indicating if the prediction $\F(\x')$ differs from $\F(\x)$. Attacks in NLP normally needs another set of Boolean functions ${C_1...C_n}$ indicating whether the perturbation satisfies a certain set of language constraints.

Initial studies on NLP adversarial attacks performed character-level perturbations to create misspellings \cite{ebrahimi2017hotflip, Gao2018BlackBoxGO, pruthi2019combating, li2018textbugger}. Recent later works have explored various word substitutions methods to produce adversarial examples in natural language. Both \citet{alzantot2018generating} and \citet{Jin2019TextFooler} use counter-fitted word embeddings to find synonyms while \citet{pwws-ren-etal-2019-generating} and \citet{pso-zang-etal-2020-word} use lexical databases like WordNet \cite{wordnet} and HowNet \cite{HowNet-2010}. Lately, masked language models have been used to perform word substitutions to preserve fluency of the perturbed text better \cite{li-etal-2020-bert-attack, garg-ramakrishnan-2020-bae, shi-huang-2020-robustness}.

%% file: 11-details-search.tex
\subsection{Details on Search for Words to Perturb}
\label{ref:search}

Solving \eref{eq:adv} needs the Chinese adversarial attacks to conduct a combinatorial search task and adapt search algorithms from the English adversarial attacks in this paper. 
 The search algorithm aims to perturb a text input with language transformations such as  synonym substitutions in order to fool a target NLP model while the perturbation adheres to  linguistic constraints.

The potential search space is exponential by nature. Assuming $\x$ includes $n$ words, and each word has $S$ potential substitutions, the total number of possible perturbed inputs is then  $(S + 1)^n - 1$. The search space of all potential adversarial examples for a given $\x$ is far too large for an exhaustive search. This is why many heuristic search algorithms were proposed in the literature, including greedy method with \importanceRankingName \cite{ Gao2018BlackBoxGO,Jin2019TextFooler, pwws-ren-etal-2019-generating}, beam search \cite{Ebrahimi2017HotFlipWA}, and population based genetic algorithm \cite{alzantot2018generating}. While heuristic search algorithms cannot guarantee an optimal solution, they can efficiently search for a valid adversarial example.

\begin{algorithm}[t]
\caption{Word Substitution Attack against Chinese NLP Models }
\label{attack-algo}
\begin{algorithmic}[1]
    \STATE {\bfseries Input:} Input text $\x$
    \STATE $\mathbf{x} = segment(\x) = [x_1, \dots, x_n]$
    \STATE $R = \;$ranking $r_1, \dots, r_n$ of words $x_1, \dots x_n$
    \STATE $\x^* = \x$
    \FOR {$i=r_1, \dots, r_n$}
        \STATE $X_{\text{candidate}} = T_1(\x, i) \cup \dots \cup T_k(\x, i) $ 
        \STATE $X_{\text{candidate}}' = \{x'|C_j(x', x) \; \forall C_j \in \{C_1...C_n\}$
        \IF{$X_{\text{candidate}}' \neq \emptyset$}
            \STATE $\x^* = \text{arg max}_{\x' \in X_{\text{candidate}}'} score(\x')$
            \IF {$F(\x^*) \neq F(\x)$}
                \STATE return $\x^*$
            \ENDIF
        \ELSE
            \STATE end search
        \ENDIF
    \ENDFOR
\end{algorithmic}
\end{algorithm}

%% file: 12-more-c.tex
\subsection{Details on Language Constraints}
\label{sec:method:morec}
NLP adversarial attacks generate perturbations and use a set of constraints to filter out undesirable $\x'$ to ensure that perturbed $\x'$ preserves the semantics and fluency of the original $\x$ \cite{Morris2020TextAttackAF}.  Therefore, we propose to use three following constraints: 
\begin{itemize}
    \item \textbf{Stop word modification}: Replacing the coordinating conjunctions and pronouns within a sentence often changes the semantics of a target sentence. Therefore, words such as "but" and "I" cannot be perturbed.
    \item \textbf{Repeat modification}: This prevents replaced words to be modified again, as the targeted word may gradually diverge from its original meaning.
    \item \textbf{Multilingual Universal Sentence Encoder (MUSE)}:\ Using the multilingual sentence encoder, we encode both original $x$ and $x'$ and measure the cosine similarity between the two text. We require that the cosine similarity is above $0.9$. 
\end{itemize}

%% file: 30-tables.tex
\begin{table}[h!]
\resizebox{0.5\textwidth}{!}{
\centering
{
\begin{tabular}{|c|c|c|c|}
\hline
\textbf{Attack Method}                                                                   & \textbf{Constraints} & \textbf{Success Rate} & \textbf{\% Perturbed} \\ \hline

\multirow{2}{*}{Open HowNet}                                                              & STM-RM         & 81.20                 & 32.54                 \\ \cline{2-4} 
                                                                                   & STM-RM-MUSE      & 56.49                  & 27.99                 \\ \hline
\multirow{2}{*}{MLM}                                                            & STM-RM         & 53.67                  & 40.33                  \\ \cline{2-4} 
                                                                                   & STM-RM-MUSE      & 48.03                  & 28.71                   \\ \hline
\multirow{2}{*}{Homonym}                                                            & STM-RM         & 74.14                  & 54.67                  \\ \cline{2-4} 
                                                                                   & STM-RM-MUSE      & 49.57                  & 38.22                   \\ \hline
\multirow{2}{*}{Morphonym}                                                            & STM-RM         & 43.21                  & 45.73                  \\ \cline{2-4} 
                                                                                   & STM-RM-MUSE      & 31.02                  & 36.63                   \\ \hline
\multirow{2}{*}{Composite}                                                            & STM-RM         & 96.00                  & 41.05                  \\ \cline{2-4} 
                                                                                   & STM-RM-MUSE      & 87.50                  & 31.74                   \\ \hline                                       
\end{tabular}
}}
\caption{Attack results of classification task performed on online-shopping review dataset. Attack success rate and amount of perturbations of each attack. ``STM-RM" stands for stop word modification and repeat modification, and ``STM-RM-MUSE" stands for stop word modification, repeat modification, and universal sentence encoder constraint.}
\label{tab:attack-result-class}
\vspace{-4mm}
\end{table}

\begin{table}[h!]
\resizebox{0.5\textwidth}{!}{
\centering
{
\begin{tabular}{|c|c|c|c|}
\hline
\textbf{Attack Method}                                                                   & \textbf{Constraints} & \textbf{Success Rate} & \textbf{\% Perturbed} \\ \hline

\multirow{2}{*}{Open HowNet}                                                              & STM-RM         & 74.94                 & 37.35                 \\ \cline{2-4} 
                                                                                   & STM-RM-MUSE      & 75.00                  & 36.59                 \\ \hline
\multirow{2}{*}{MLM}                                                            & STM-RM         & 75.40                  & 40.88                  \\ \cline{2-4} 
                                                                                   & STM-RM-MUSE      & 75.79                 & 40.93                   \\ \hline
\multirow{2}{*}{Homonym}                                                            & STM-RM         & 86.21                  & 49.63                  \\ \cline{2-4} 
                                                                                   & STM-RM-MUSE      & 86.21                  & 49.58                   \\ \hline
\multirow{2}{*}{Morphonym}                                                            & STM-RM         & 60.23                  & 46.47                  \\ \cline{2-4} 
                                                                                   & STM-RM-MUSE      & 59.81                  & 46.60                   \\ \hline
\multirow{2}{*}{Composite}                                                            & STM-RM         & 98.16                 & 39.06                  \\ \cline{2-4} 
                                                                                   & STM-RM-MUSE      & 98.02                  & 38.74                   \\ \hline                      
\end{tabular}
}}
\caption{Attack results on Chinese entailment model using the Chinanews dataset. Attacks' setup same as \tref{tab:attack-result-class}.}
\label{tab:attack-result-entail}
\vspace{-4mm}
\end{table}

\begin{table}[h!]
\resizebox{0.5\textwidth}{!}{
\centering
{
\begin{tabular}{|c|c|c|c|}
\hline
\textbf{Attack Method}                                                                   & \textbf{Consistency(1-5)} & \textbf{Fluency(1-5)} & \textbf{$\Delta$ Fluency} \\ \hline
{Open HowNet}                                                              & 2.94        & 2.69                 & -2.31                                                \\ \hline
{MLM}                                                             & 3.41        & 3.70                  & -1.28                                \\ \hline
{Homonym}                                                               & 4.44         & 4.31                  & -0.69                                            \\ \hline
{Morphonym}                                                           & 3.75         & 3.31                  & -1.69                          \\ \hline
{Composite}             & 3.13         & 2.94                  & -2.06                                                                     \\ \hline                                                                        
\end{tabular}
}}
\caption{Human evaluation of attacks on Online-shopping dataset. We report average consistency and fluency scores on examples generated from each attack method. STM-RM-MUSE constraints were used for all attack methods.}
\label{fig:human_eval_result_class}
\end{table}

\begin{table}[h!]
\resizebox{0.5\textwidth}{!}{
\centering
{
\begin{tabular}{|c|c|c|c|}
\hline
\textbf{Attack Method}                                                                   & \textbf{Consistency(1-5)} & \textbf{Fluency(1-5)} & \textbf{$\Delta$ Fluency} \\ \hline

{Open HowNet}                                                              & 3.10        & 3.07                 & -1.93                                                \\ \hline
{MLM}                                                             & 3.20        & 3.90                  & -0.87                                \\ \hline
{Homonym}                                                               & 4.43         & 3.57                  & -1.33                                            \\ \hline
{Morphonym}                                                           & 3.90         & 3.73                  & -1.27                          \\ \hline
{Composite}             & 2.97         & 3.43                  & -2.53                                                                     \\ \hline                                                                        
\end{tabular}
}}
\caption{Human evaluation of attacks on Chinanews dataset. STM-RM-MUSE constraints were used for all attack methods.}
\label{fig:human_eval_result_entail}
\vspace{-5mm}
\end{table}

%% file: 22-at-results.tex
\subsection{Adversarial Training}
\label{sec:at}

\begin{table}[h!]
\resizebox{0.5\textwidth}{!}{
\centering
{
\begin{tabular}{|c|c|c|c|c|}
\hline
\textbf{Attack Method}                                                                   & \textbf{Constraints} & \textbf{Pre Success Rate} & \textbf{AT Success Rate} & \textbf{$\Delta\%$} \\ \hline

\multirow{2}{*}{Open HowNet}                                                              & STM-RM         & 81.20                 & 65.05    & -19.89             \\ \cline{2-5} 
                                                                                   & STM-RM-MUSE      & 56.49                  & 43.30   &-23.35              \\ \hline
\multirow{2}{*}{MLM}                                                            & STM-RM         & 53.67                  & 36.10 & -32.74                 \\ \cline{2-5} 
                                                                                   & STM-RM-MUSE      & 48.03                 & 38.94 &-18.93                   \\ \hline
\multirow{2}{*}{Homonym}                                                            & STM-RM         & 74.14                  & 40.83 & -44.93                  \\ \cline{2-5} 
                                                                                   & STM-RM-MUSE      & 49.57                  & 35.86 &-27.66                   \\ \hline
\multirow{2}{*}{Morphonym}                                                            & STM-RM         & 43.21                  & 30.51 &-29.39                  \\ \cline{2-5} 
                                                                                   & STM-RM-MUSE      & 31.02                  & 15.72 & -49.32                   \\ \hline
\multirow{2}{*}{Composite}                                                            & STM-RM         & 96.00	                & 77.75 & -19.01                 \\ \cline{2-5} 
                                                                                   & STM-RM-MUSE      & 87.50 &	60.34 & -31.04               \\ \hline                      
\end{tabular}
}}
\caption{Results of adversarial training performed on BERT model. "Pre Success Rate" stands for the success rate of composite attack on the pre-adversarial-trained model, and "AT Success Rate" stands for the success rate of composite attack on adversarial-trained model }
\label{fig:at-result-class}
\end{table}

To evaluate how adversarial examples generated by the attack methods could improve the adversarial robustness of a target model, we attacked the target BERT model (Chinese sentiment classification) with 1000 examples from the training set of an online shopping review dataset, and finetune the target model with the successfully attacked examples. The model was trained for 3 epochs with 1 initial clean epochs,  learning rate of 5e-5, and effective batch size of 32 (8x4).

\tref{fig:at-result-class} shows the positive effect of adversarial training (AT) that improve the robustness of Chinese language models against all five of our proposed attacks. For instance, on the target BERT model, attack success rate decreased by up to 49.32$\%$ after being trained by adversarial examples generated by the Composite-MUSE attack method. Across all attacks, the AT-trained models result with a significant drop in attack success rate.

%% file: 20-more-data.tex
\subsection{Victim Model and Dataset Setup}
\label{sec:data}

We chose to perform attacks on Chinese sentiment classification and entailment models, and chose BERT and RoBERTa as our victim models due to their reported robustness against perturbations when compared to other models such as LSTM and CNN \cite{hsieh-etal-2019-robustness}.

 The target BERT model for Chinese sentiment classification is from Huggingface \footnote{\url{https://huggingface.co/Raychanan/bert-base-chinese-FineTuned-Binary-Best}}, and the target RoBERTa model for entailment was trained on the training set of the Chinanews dataset \cite{glyph-project}. The validation performance of the BERT sentiment classification model  is $89.80\%$ and is $89.71\%$ for RoBERTa entailment model.

For both models, untargeted classification was set as the fooling goal function and the search method was greedy search with word-importance-ranking as aforementioned. 500 examples were attacked by using each attack method. Two related datasets are as follows.

\begin{itemize}[leftmargin=*]
\item Dataset-1: An online shopping review dataset\footnote{\url{https://github.com/SophonPlus/ChineseNlpCorpus/tree/master/datasets/online_shopping_10_cats}} for sentiment classification tasks was used to generate attacks against the BERT classification model, with 500 examples from the test set to check the model's adversarial robustness. 

\item Dataset-2: The Chinanews dataset was collected by the glyph project \cite{glyph-project} and consists of the summary and first paragraphs of news articles from chinanews.com. Each set of summary and first paragraph was labeled with one of 7 topic classes, including mainland China politics, Hong Kong-Macau politics, Taiwan politics, International news,
financial news, culture, entertainment, sports, and health. We randomly sampled 500 examples from the test set to attack against the entailment model.
\end{itemize}

%% file: 23-discussion.tex
\subsection{Discussion of Results}
\label{sec:discuss}
After checking the generated adversarial examples, we realize that a leading cause behind inconsistent and unnatural adversarial examples for Open HowNet transformation is out-of-context substitutions, supported by it having the highest attack success rate yet the lowest consistency/fluency score. Most models are sufficiently robust to attacks with common synonyms, which means successful attacks are often accomplished by distant and unconventional synonym substitutions. On the other hand, cases of out-of-context word substitutions were observed less often in the other attack methods. This is reasonable as homonym and morphonym attack methods only perturb the presentation of the substituted words without changing its semantics to human, while a classification and entailment models fail to attend to the context.  However, in rare cases, homonym transformations are also prone to out-of-context substitutions as some Chinese characters have multiple pronunciations. In such scenarios, homonym attacks may result in a false successful attack due to failures to recognize the correct pronunciation and provide an appropriate substitution.

Furthermore, we also observe that perturbing certain characters results in almost guaranteed change in prediction, which was first reported by \citet{wang2020evaluatingChineseBERT}. For instance, the Chinese character "bu" translates to "no" in English. As illustrated by the first example in ~\fref{fig:examples1}d, when "bu" is replaced by its morphonym or homonym, the prediction of the perturbed sentence often changes from negative to positive, as a strong negative cue was replaced by another character that the victim model not yet recognizes. Similarly, in the case of entailment models, when the name of a country/region is substituted with its morphonym or homonym, examples with region-specific labels (Hong kong-macau politic, Mainland china politics, etc.) were most often attacked successfully. The vulnerability of Chinese BERT and RoBERTa models against morphonym and homonym adversarial attacks indicates that there is still a large room for improvement in their adversarial robustness.

%% file: 25-conclusion.tex
\section{Conclusion}
\label{sec:conclusion}

In summary, we investigate how to adapt SOTA adversarial attack algorithms to the Chinese language. Our experiments show that the system of generating English adversarial examples can be sufficiently adapted to Chinese, given appropriate text segmentation, perturbation methods, and linguistic constraints. We also introduce two additional perturbation methods particular to the attributes of the Chinese language. Because most of the English/Chinese-specific components of the workflow can be substituted with other languages and resources, we are optimistic that the adaptation workflow presented in this paper can be generalized to other languages in building a language-agnostic attack algorithm in future research.

%% file: 32-figure.tex
\begin{figure*}[h!]
     \centering
     \begin{subfigure}[t]{0.5\textwidth}
         \centering
         \includegraphics[width=\textwidth]{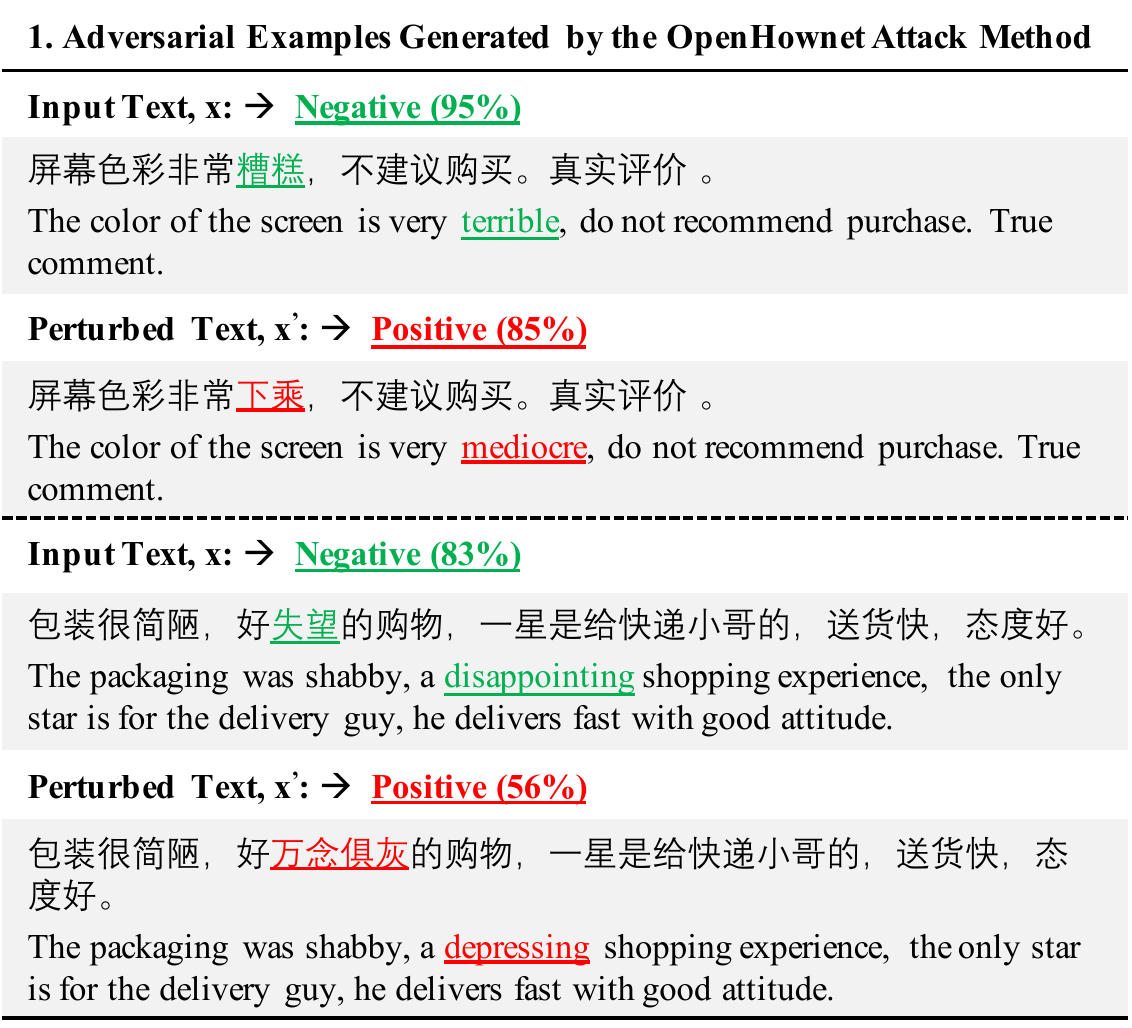}
 \caption{Adversarial examples of Open HowNet}
 \label{fig:example-HowNet}
     \end{subfigure}
     \hfill
     \begin{subfigure}[t]{0.49\textwidth}
         \centering
         \includegraphics[width=\textwidth]{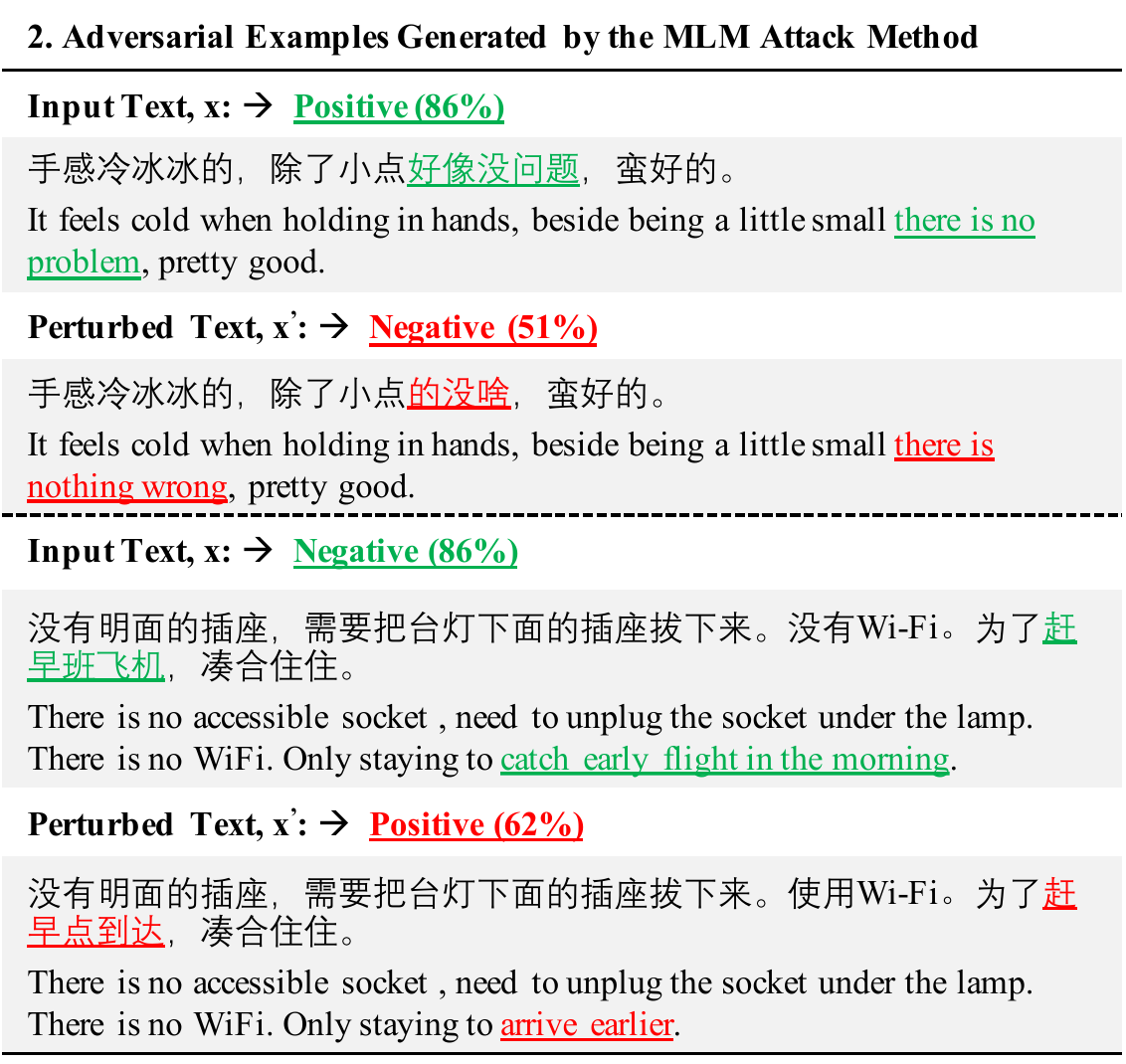}
 \caption{Adversarial examples of MLM}
 \label{fig:example-mlm}
     \end{subfigure}
     \hfill
     \begin{subfigure}[b]{0.5\textwidth}
         \centering
         \includegraphics[width=\textwidth]{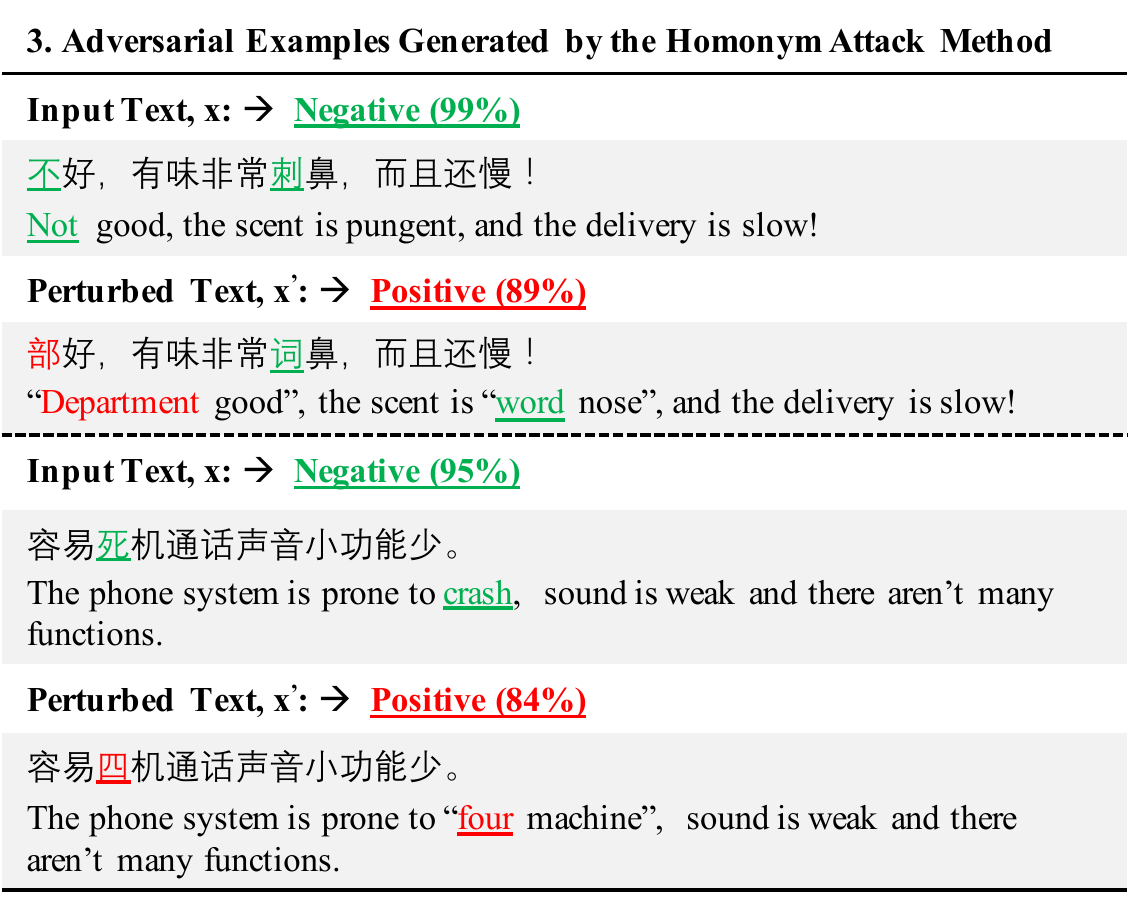}
 \caption{Adversarial examples of Homonym}
 \label{fig:example-homonym}
     \end{subfigure}
     \hfill
     \begin{subfigure}[b]{0.49\textwidth}
         \centering
         \includegraphics[width=\textwidth]{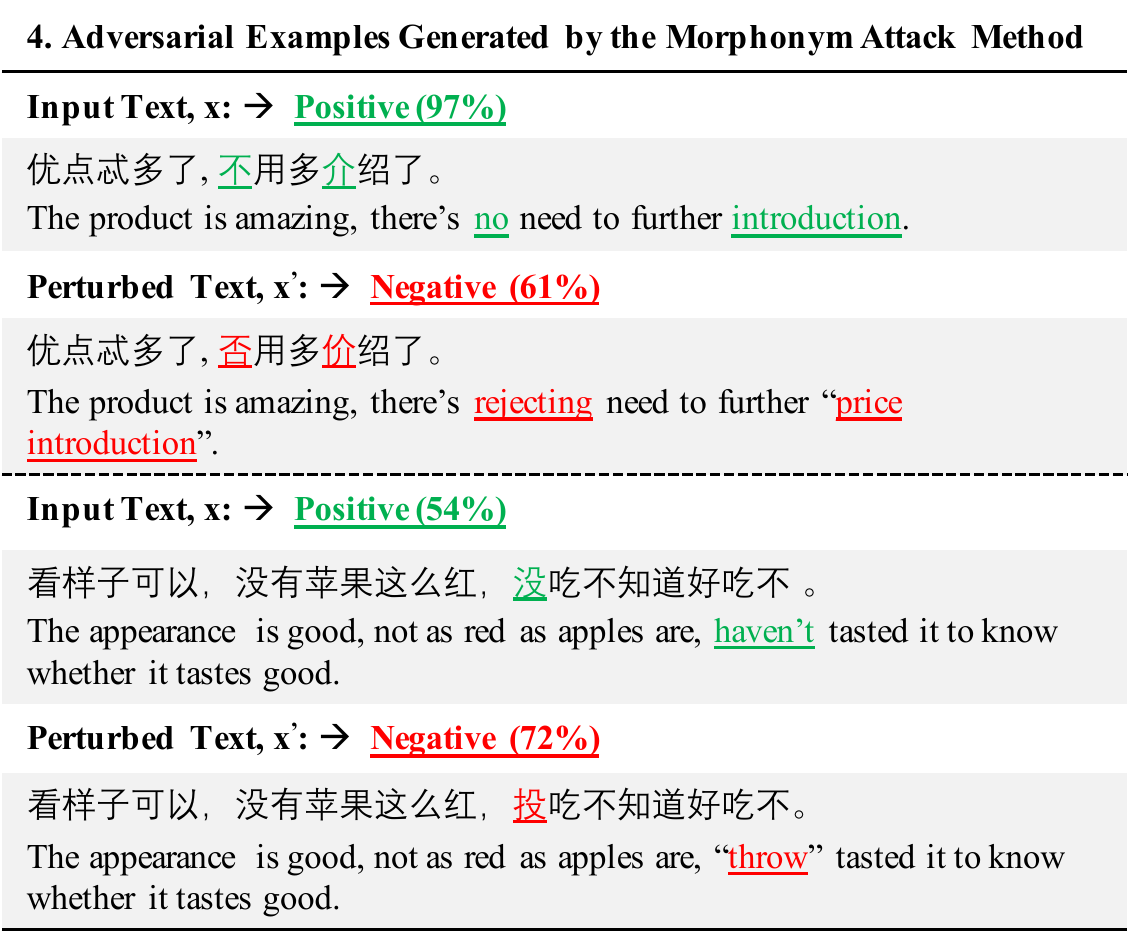}
 \caption{Adversarial examples of Morphonym}
 \label{fig:example-morphonym}
     \end{subfigure}
        \caption{Selected Adversarial Examples generated by proposed adversarial attacks on the online shopping review dataset (classification). Note for adversarial examples generated by the Homonym and Morphonym attack method (figure 2c and 2d), word substitutions are based on Chinese language characteristics instead of semantic meaning. For examples in figure 2c, substitutions were chosen from characters with similar sounds. For figure 2d, substitutions were from characters that look similar to human readers. }
        \label{fig:examples1}
\end{figure*}

%% file: 31-figure.tex
\begin{figure*}[h!]
     \centering
     \begin{subfigure}[t]{0.5\textwidth}
         \centering
         \includegraphics[width=\textwidth]{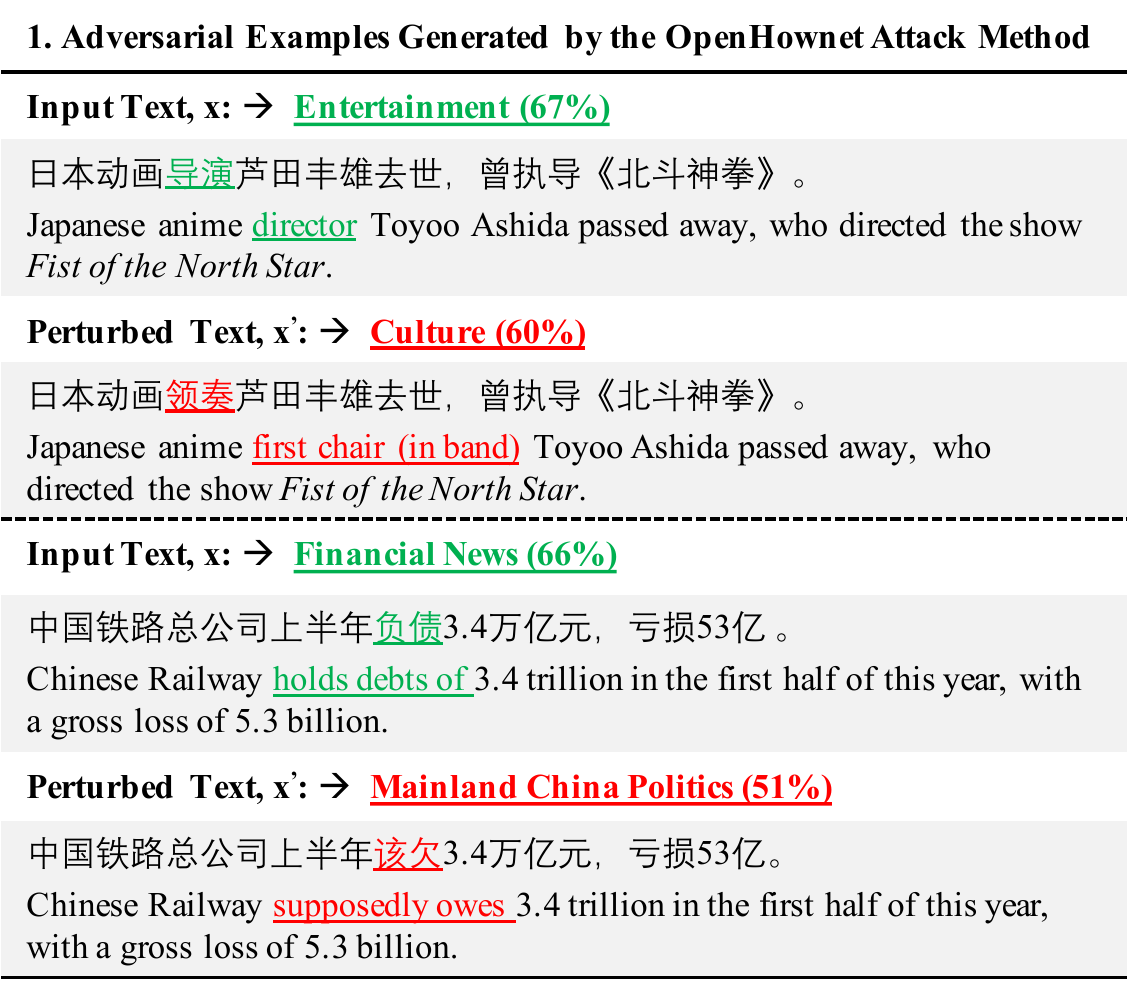}
 \caption{Adversarial examples of Open HowNet}
 \label{fig:example-HowNet}
     \end{subfigure}
     \hfill
     \begin{subfigure}[t]{0.49\textwidth}
         \centering
         \includegraphics[width=\textwidth]{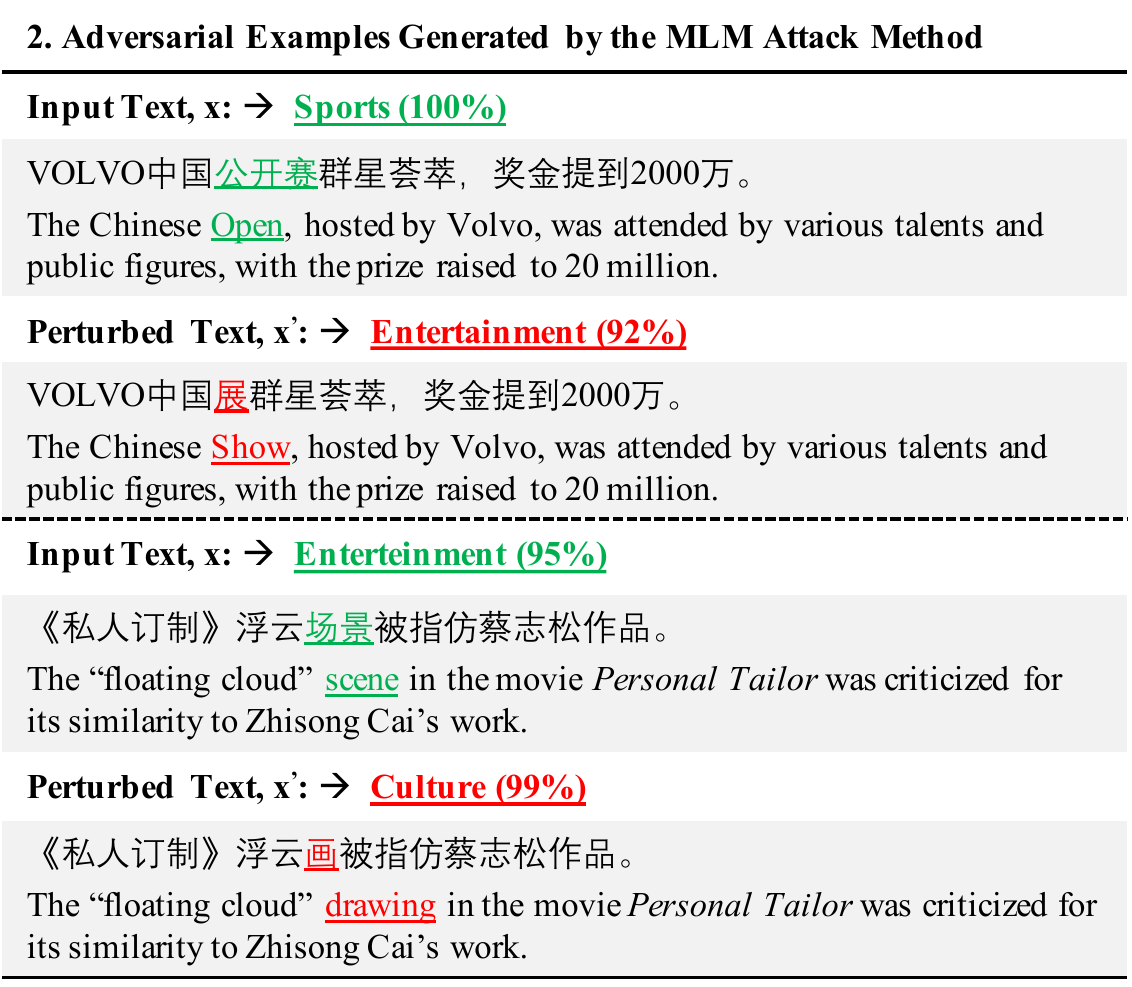}
 \caption{Adversarial examples of MLM}
 \label{fig:example-mlm}
     \end{subfigure}
     \hfill
     \begin{subfigure}[b]{0.5\textwidth}
         \centering
         \includegraphics[width=\textwidth]{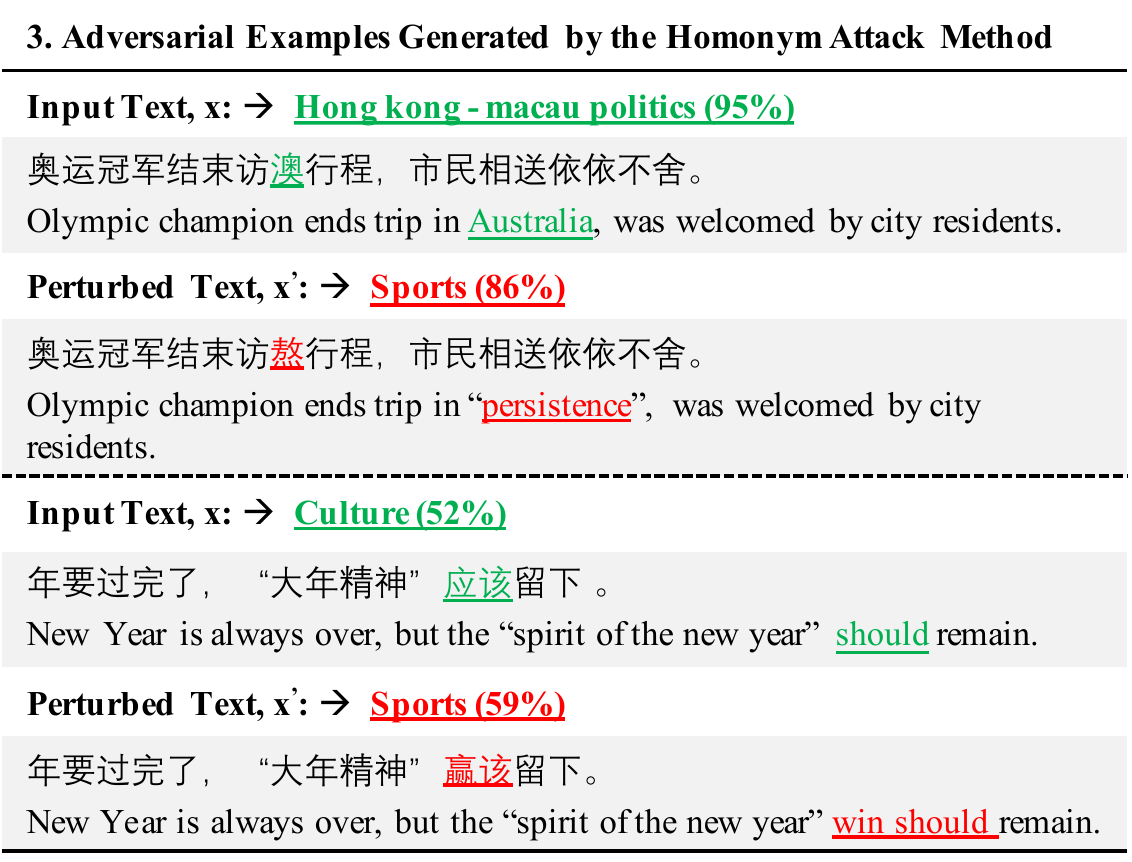}
 \caption{Adversarial examples of Homonym}
 \label{fig:example-homonym}
     \end{subfigure}
     \hfill
     \begin{subfigure}[b]{0.49\textwidth}
         \centering
         \includegraphics[width=\textwidth]{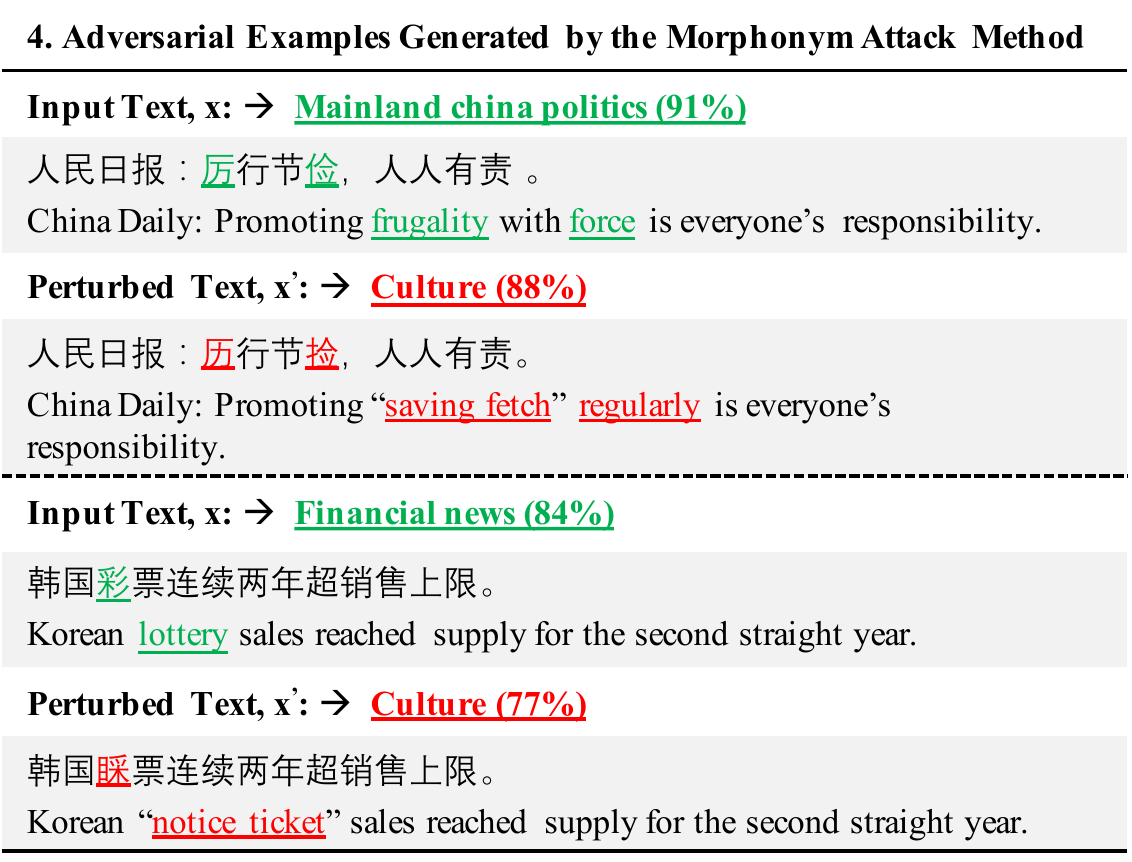}
 \caption{Adversarial examples of Morphonym}
 \label{fig:example-morphonym}
     \end{subfigure}
        \caption{Selected Adversarial Examples generated by proposed adversarial attacks on the Chinanews dataset (entailment task). As mentioned in the discussion section, substituting specific characters almost guarantees a change in the prediction result. As shown by the second example in figure 3c, the homonym substitution of the word "should" added the new semantic meaning of "winning", which is a strong cue for the Sports category.}
        \label{fig:examples2}
\end{figure*}

%% file: 00-aclwkp2023-4page.bbl
\begin{thebibliography}{33}
\expandafter\ifx\csname natexlab\endcsname\relax\def\natexlab#1{#1}\fi

\bibitem[{Alzantot et~al.(2018)Alzantot, Sharma, Elgohary, Ho, Srivastava, and
  Chang}]{alzantot2018generating}
Moustafa Alzantot, Yash Sharma, Ahmed Elgohary, Bo-Jhang Ho, Mani Srivastava,
  and Kai-Wei Chang. 2018.
\newblock Generating natural language adversarial examples.
\newblock \emph{arXiv preprint arXiv:1804.07998}.

\bibitem[{Cer et~al.(2018)Cer, Yang, Kong, Hua, Limtiaco, John, Constant,
  Guajardo{-}Cespedes, Yuan, Tar, Sung, Strope, and Kurzweil}]{Cer18USE}
Daniel Cer, Yinfei Yang, Sheng{-}yi Kong, Nan Hua, Nicole Limtiaco, Rhomni~St.
  John, Noah Constant, Mario Guajardo{-}Cespedes, Steve Yuan, Chris Tar,
  Yun{-}Hsuan Sung, Brian Strope, and Ray Kurzweil. 2018.
\newblock \href {http://arxiv.org/abs/1803.11175} {Universal sentence encoder}.
\newblock \emph{CoRR}, abs/1803.11175.

\bibitem[{Conneau et~al.(2019)Conneau, Khandelwal, Goyal, Chaudhary, Wenzek,
  Guzmán, Grave, Ott, Zettlemoyer, and Stoyanov}]{xlm-roberta}
Alexis Conneau, Kartikay Khandelwal, Naman Goyal, Vishrav Chaudhary, Guillaume
  Wenzek, Francisco Guzmán, Edouard Grave, Myle Ott, Luke Zettlemoyer, and
  Veselin Stoyanov. 2019.
\newblock \href {https://doi.org/10.48550/ARXIV.1911.02116} {Unsupervised
  cross-lingual representation learning at scale}.

\bibitem[{Devlin et~al.(2018{\natexlab{a}})Devlin, Chang, Lee, and
  Toutanova}]{devlin2018bert}
Jacob Devlin, Ming-Wei Chang, Kenton Lee, and Kristina Toutanova.
  2018{\natexlab{a}}.
\newblock Bert: Pre-training of deep bidirectional transformers for language
  understanding.
\newblock \emph{arXiv preprint arXiv:1810.04805}.

\bibitem[{Devlin et~al.(2018{\natexlab{b}})Devlin, Chang, Lee, and
  Toutanova}]{Devlin2018BERTPO}
Jacob Devlin, Ming-Wei Chang, Kenton Lee, and Kristina Toutanova.
  2018{\natexlab{b}}.
\newblock Bert: Pre-training of deep bidirectional transformers for language
  understanding.
\newblock In \emph{NAACL-HLT}.

\bibitem[{Dong et~al.(2010)Dong, Dong, and Hao}]{HowNet-2010}
Zhendong Dong, Qiang Dong, and Changling Hao. 2010.
\newblock Hownet and its computation of meaning.
\newblock In \emph{Proceedings of the 23rd International Conference on
  Computational Linguistics: Demonstrations}, COLING ’10, page 53–56, USA.
  Association for Computational Linguistics.

\bibitem[{Ebrahimi et~al.(2017{\natexlab{a}})Ebrahimi, Rao, Lowd, and
  Dou}]{ebrahimi2017hotflip}
Javid Ebrahimi, Anyi Rao, Daniel Lowd, and Dejing Dou. 2017{\natexlab{a}}.
\newblock Hotflip: White-box adversarial examples for text classification.
\newblock \emph{arXiv preprint arXiv:1712.06751}.

\bibitem[{Ebrahimi et~al.(2017{\natexlab{b}})Ebrahimi, Rao, Lowd, and
  Dou}]{Ebrahimi2017HotFlipWA}
Javid Ebrahimi, Anyi Rao, Daniel Lowd, and Dejing Dou. 2017{\natexlab{b}}.
\newblock Hotflip: White-box adversarial examples for text classification.
\newblock In \emph{ACL}.

\bibitem[{Gao et~al.(2018)Gao, Lanchantin, Soffa, and Qi}]{Gao2018BlackBoxGO}
Ji~Gao, Jack Lanchantin, Mary~Lou Soffa, and Yanjun Qi. 2018.
\newblock Black-box generation of adversarial text sequences to evade deep
  learning classifiers.
\newblock \emph{2018 IEEE Security and Privacy Workshops (SPW)}, pages 50--56.

\bibitem[{Garg and Ramakrishnan(2020)}]{garg-ramakrishnan-2020-bae}
Siddhant Garg and Goutham Ramakrishnan. 2020.
\newblock \href {https://doi.org/10.18653/v1/2020.emnlp-main.498} {{BAE}:
  {BERT}-based adversarial examples for text classification}.
\newblock In \emph{Proceedings of the 2020 Conference on Empirical Methods in
  Natural Language Processing (EMNLP)}, pages 6174--6181, Online. Association
  for Computational Linguistics.

\bibitem[{Goodfellow et~al.(2014)Goodfellow, Shlens, and
  Szegedy}]{goodfellow2014explaining}
Ian~J Goodfellow, Jonathon Shlens, and Christian Szegedy. 2014.
\newblock Explaining and harnessing adversarial examples.
\newblock \emph{arXiv preprint arXiv:1412.6572}.

\bibitem[{Hsieh et~al.(2019)Hsieh, Cheng, Juan, Wei, Hsu, and
  Hsieh}]{hsieh-etal-2019-robustness}
Yu-Lun Hsieh, Minhao Cheng, Da-Cheng Juan, Wei Wei, Wen-Lian Hsu, and Cho-Jui
  Hsieh. 2019.
\newblock \href {https://doi.org/10.18653/v1/P19-1147} {On the robustness of
  self-attentive models}.
\newblock In \emph{Proceedings of the 57th Annual Meeting of the Association
  for Computational Linguistics}, pages 1520--1529, Florence, Italy.
  Association for Computational Linguistics.

\bibitem[{Jin et~al.(2019)Jin, Jin, Zhou, and Szolovits}]{Jin2019TextFooler}
Di~Jin, Zhijing Jin, Joey~Tianyi Zhou, and Peter Szolovits. 2019.
\newblock Is bert really robust? natural language attack on text classification
  and entailment.
\newblock \emph{ArXiv}, abs/1907.11932.

\bibitem[{Li et~al.(2020{\natexlab{a}})Li, Du, Ji, Zhang, Lu, Yang, and
  Wang}]{textshield}
Jinfeng Li, Tianyu Du, Shouling Ji, Rong Zhang, Quan Lu, Min Yang, and Ting
  Wang. 2020{\natexlab{a}}.
\newblock Textshield: Robust text classification based on multimodal embedding
  and neural machine translation.
\newblock In \emph{USENIX Security Symposium}.

\bibitem[{Li et~al.(2018)Li, Ji, Du, Li, and Wang}]{li2018textbugger}
Jinfeng Li, Shouling Ji, Tianyu Du, Bo~Li, and Ting Wang. 2018.
\newblock Textbugger: Generating adversarial text against real-world
  applications.
\newblock \emph{arXiv preprint arXiv:1812.05271}.

\bibitem[{Li et~al.(2020{\natexlab{b}})Li, Ma, Guo, Xue, and
  Qiu}]{li-etal-2020-bert-attack}
Linyang Li, Ruotian Ma, Qipeng Guo, Xiangyang Xue, and Xipeng Qiu.
  2020{\natexlab{b}}.
\newblock \href {https://doi.org/10.18653/v1/2020.emnlp-main.500}
  {{BERT}-{ATTACK}: Adversarial attack against {BERT} using {BERT}}.
\newblock In \emph{Proceedings of the 2020 Conference on Empirical Methods in
  Natural Language Processing (EMNLP)}, pages 6193--6202, Online. Association
  for Computational Linguistics.

\bibitem[{Miller(1995)}]{wordnet}
George~A. Miller. 1995.
\newblock \href {https://doi.org/10.1145/219717.219748} {Wordnet: A lexical
  database for english}.
\newblock \emph{Commun. ACM}, 38(11):39–41.

\bibitem[{Moosavi-Dezfooli et~al.(2016)Moosavi-Dezfooli, Fawzi, and
  Frossard}]{moosavi2016deepfool}
Seyed-Mohsen Moosavi-Dezfooli, Alhussein Fawzi, and Pascal Frossard. 2016.
\newblock {DeepFool}: a simple and accurate method to fool deep neural
  networks.
\newblock In \emph{IEEE Conference on Computer Vision and Pattern Recognition
  (CVPR)}.

\bibitem[{Morris et~al.(2020{\natexlab{a}})Morris, Lifland, Yoo, and
  Qi}]{Morris2020TextAttackAF}
John Morris, Eli Lifland, Jin~Yong Yoo, and Yanjun Qi. 2020{\natexlab{a}}.
\newblock {TextAttack}: A framework for adversarial attacks in natural language
  processing.
\newblock \emph{ArXiv}, abs/2005.05909.

\bibitem[{Morris et~al.(2020{\natexlab{b}})Morris, Lifland, Lanchantin, Ji, and
  Qi}]{morris2020reevaluating}
John~X. Morris, Eli Lifland, Jack Lanchantin, Yangfeng Ji, and Yanjun Qi.
  2020{\natexlab{b}}.
\newblock \href {http://arxiv.org/abs/2004.14174} {Reevaluating adversarial
  examples in natural language}.

\bibitem[{Papernot et~al.(2016)Papernot, McDaniel, Jha, Fredrikson, Celik, and
  Swami}]{papernot2016limitations}
Nicolas Papernot, Patrick McDaniel, Somesh Jha, Matt Fredrikson, Z~Berkay
  Celik, and Ananthram Swami. 2016.
\newblock The limitations of deep learning in adversarial settings.
\newblock In \emph{IEEE European Symposium on Security and Privacy (EuroS\&P)}.

\bibitem[{Pruthi et~al.(2019)Pruthi, Dhingra, and Lipton}]{pruthi2019combating}
Danish Pruthi, Bhuwan Dhingra, and Zachary~C Lipton. 2019.
\newblock Combating adversarial misspellings with robust word recognition.
\newblock \emph{arXiv preprint arXiv:1905.11268}.

\bibitem[{Qi et~al.(2019)Qi, Yang, Liu, Dong, Sun, and Dong}]{qi2019OpenHowNet}
Fanchao Qi, Chenghao Yang, Zhiyuan Liu, Qiang Dong, Maosong Sun, and Zhendong
  Dong. 2019.
\newblock Openhownet: An open sememe-based lexical knowledge base.
\newblock \emph{arXiv preprint arXiv:1901.09957}.

\bibitem[{Ren et~al.(2019)Ren, Deng, He, and
  Che}]{pwws-ren-etal-2019-generating}
Shuhuai Ren, Yihe Deng, Kun He, and Wanxiang Che. 2019.
\newblock \href {https://doi.org/10.18653/v1/P19-1103} {Generating natural
  language adversarial examples through probability weighted word saliency}.
\newblock In \emph{Proceedings of the 57th Annual Meeting of the Association
  for Computational Linguistics}, pages 1085--1097, Florence, Italy.
  Association for Computational Linguistics.

\bibitem[{Shao and Wang(2022)}]{GPSAttack}
Yuyao Shao and Liming Wang. 2022.
\newblock \href {https://doi.org/10.1109/IJCNN55064.2022.9892804} {Gpsattack: A
  unified glyphs, phonetics and semantics multi-modal attack against chinese
  text classification models}.
\newblock In \emph{2022 International Joint Conference on Neural Networks
  (IJCNN)}, pages 1--8.

\bibitem[{Shi and Huang(2020)}]{shi-huang-2020-robustness}
Zhouxing Shi and Minlie Huang. 2020.
\newblock \href {https://doi.org/10.18653/v1/2020.findings-emnlp.16}
  {Robustness to modification with shared words in paraphrase identification}.
\newblock In \emph{Findings of the Association for Computational Linguistics:
  EMNLP 2020}, pages 164--171, Online. Association for Computational
  Linguistics.

\bibitem[{Szegedy et~al.(2013)Szegedy, Zaremba, Sutskever, Bruna, Erhan,
  Goodfellow, and Fergus}]{szegedy2013intriguing}
Christian Szegedy, Wojciech Zaremba, Ilya Sutskever, Joan Bruna, Dumitru Erhan,
  Ian Goodfellow, and Rob Fergus. 2013.
\newblock \href {http://arxiv.org/abs/1312.6199} {Intriguing properties of
  neural networks}.
\newblock \emph{arXiv preprint arXiv:1312.6199}.

\bibitem[{Wang et~al.(2020)Wang, Pan, Li, and
  Li}]{wang2020evaluatingChineseBERT}
Boxin Wang, Boyuan Pan, Xin Li, and Bo~Li. 2020.
\newblock \href {http://arxiv.org/abs/2004.03742} {Towards evaluating the
  robustness of chinese bert classifiers}.

\bibitem[{Wang et~al.(2022)Wang, Xu, Liu, Cheng, and
  Li}]{wang-etal-2022-semattack}
Boxin Wang, Chejian Xu, Xiangyu Liu, Yu~Cheng, and Bo~Li. 2022.
\newblock \href {https://doi.org/10.18653/v1/2022.findings-naacl.14}
  {{S}em{A}ttack: Natural textual attacks via different semantic spaces}.
\newblock In \emph{Findings of the Association for Computational Linguistics:
  NAACL 2022}, pages 176--205, Seattle, United States. Association for
  Computational Linguistics.

\bibitem[{Zang et~al.(2020)Zang, Qi, Yang, Liu, Zhang, Liu, and
  Sun}]{pso-zang-etal-2020-word}
Yuan Zang, Fanchao Qi, Chenghao Yang, Zhiyuan Liu, Meng Zhang, Qun Liu, and
  Maosong Sun. 2020.
\newblock \href {https://www.aclweb.org/anthology/2020.acl-main.540}
  {Word-level textual adversarial attacking as combinatorial optimization}.
\newblock In \emph{Proceedings of the 58th Annual Meeting of the Association
  for Computational Linguistics}, pages 6066--6080, Online. Association for
  Computational Linguistics.

\bibitem[{Zhang and LeCun(2017)}]{glyph-project}
Xiang Zhang and Yann LeCun. 2017.
\newblock \href {https://doi.org/10.48550/ARXIV.1708.02657} {Which encoding is
  the best for text classification in chinese, english, japanese and korean?}

\bibitem[{Zhang et~al.(2022)Zhang, Li, Shi, Yuan, Liu, Zhang, Xue, Sun, and
  Zhang}]{zhang-etal-2022-rochbert}
Zihan Zhang, Jinfeng Li, Ning Shi, Bo~Yuan, Xiangyu Liu, Rong Zhang, Hui Xue,
  Donghong Sun, and Chao Zhang. 2022.
\newblock \href {https://aclanthology.org/2022.findings-emnlp.256}
  {{R}o{C}h{B}ert: Towards robust {BERT} fine-tuning for {C}hinese}.
\newblock In \emph{Findings of the Association for Computational Linguistics:
  EMNLP 2022}, pages 3502--3516, Abu Dhabi, United Arab Emirates. Association
  for Computational Linguistics.

\bibitem[{Zhang et~al.(2020)Zhang, Liu, Zhang, Zhang, Li, Li, Duan, and
  Sun}]{argot2020Chinese}
Zihan Zhang, Mingxuan Liu, Chao Zhang, Yiming Zhang, Zhou Li, Qi~Li, Haixin
  Duan, and Donghong Sun. 2020.
\newblock \href {https://doi.org/10.24963/ijcai.2020/351} {Argot: Generating
  adversarial readable chinese texts}.
\newblock In \emph{Proceedings of the Twenty-Ninth International Joint
  Conference on Artificial Intelligence, {IJCAI-20}}, pages 2533--2539.
  International Joint Conferences on Artificial Intelligence Organization.

\end{thebibliography}
